\documentclass[11pt]{article}

\usepackage[preprint]{acl}
\usepackage{times}
\usepackage{latexsym}
\usepackage[T1]{fontenc}

\usepackage[utf8]{inputenc}

\usepackage{microtype}

\usepackage{inconsolata}

\usepackage{graphicx}

\title{CSMCIR: CoT-Enhanced Symmetric Alignment with Memory Bank for
Composed Image Retrieval}

\usepackage{fontawesome}

\author{
  \textbf{Zhipeng Qian}\textsuperscript{1*},
  \textbf{Zihan Liang}\textsuperscript{1*},
  \textbf{Yufei Ma}\textsuperscript{1*},
  \textbf{Ben Chen}\textsuperscript{1\dag},
  \textbf{Huangyu Dai}\textsuperscript{1},\\
  \textbf{Yiwei Ma}\textsuperscript{2},
  \textbf{Jiayi Ji}\textsuperscript{2},
  \textbf{Chenyi Lei}\textsuperscript{1},
  \textbf{Han Li}\textsuperscript{1},
  \textbf{Xiaoshuai Sun}\textsuperscript{2}\\
  \\
  \textsuperscript{1}Kuaishou Technology \\
  \textsuperscript{2}Key Laboratory of Multimedia Trusted Perception and Efficient Computing,\\
  \quad Ministry of Education of China, Xiamen University, 361005, P.R. China \\
  \texttt{qianzhipeng@stu.xmu.edu.cn}
  \quad \faEnvelope \quad 
  \texttt{benchen4395@gmail.com}
}
\usepackage{booktabs}
\usepackage{graphicx}
\usepackage{multirow} 
\usepackage{amsmath} 
\usepackage{booktabs}
\usepackage[switch]{lineno}
\usepackage{colortbl}
\usepackage[table]{xcolor}
\usepackage{algorithm}
\usepackage{algorithmic}
\begin{document}
\maketitle

\renewcommand{\thefootnote}{\fnsymbol{footnote}}
\footnotetext[1]{These authors contributed equally.}
\footnotetext[2]{The corresponding author.}
\renewcommand{\thefootnote}{\arabic{footnote}}
\begin{abstract}
Composed Image Retrieval (CIR) enables users to search for target images using both a reference image and manipulation text, offering substantial advantages over single-modality retrieval systems.
However, existing CIR methods suffer from representation space fragmentation: queries and targets comprise heterogeneous modalities and are processed by distinct encoders, forcing models to bridge misaligned representation spaces only through post-hoc alignment, which fundamentally limits retrieval performance. As evidenced by t-SNE visualization in Fig.~\ref{fig:tsne}(a), this architectural asymmetry manifests as three distinct, well-separated clusters in the feature space, directly demonstrating how heterogeneous modalities and architectural asymmetry  create fundamentally misaligned representation spaces from initialization.
In this work, we propose CSMCIR, a unified representation framework that achieves efficient query-target alignment through three synergistic components.
First, we introduce a Multi-level Chain-of-Thought (MCoT) prompting strategy that guides Multimodal Large Language Models to generate discriminative, semantically compatible captions for target images, establishing modal symmetry.
Building upon this, we design a symmetric dual-tower architecture where both query and target sides utilize the identical shared-parameter Q-Former for cross-modal encoding, ensuring consistent feature representations and further reducing the alignment gap.
Finally, this architectural symmetry enables an entropy-based, temporally dynamic Memory Bank strategy that provides high-quality negative samples while maintaining consistency with the evolving model state.
Extensive experiments on four datasets demonstrate that our CSMCIR achieves state-of-the-art performance with superior training efficiency. Our code is availabe at \url{https://github.com/qzp2018/CSMCIR}.

\end{abstract}
\begin{figure}[t] 
  \centering 
  \includegraphics[width=\columnwidth]{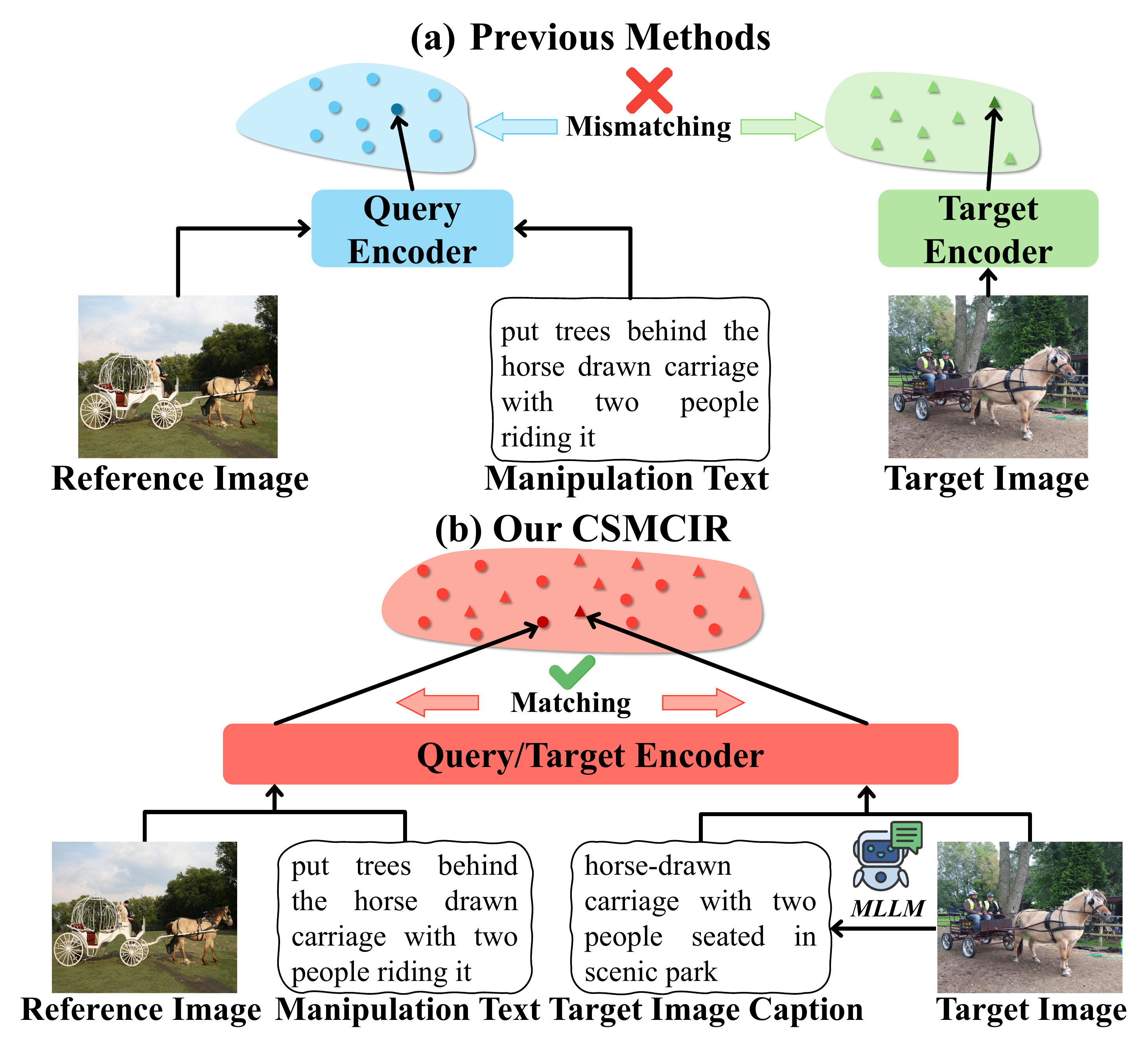
  } 
  \vspace{-0.8cm}
  \caption{Workflows of existing CIR methods (a) and our proposed CSMCIR (b). Our approach achieves modal and structural symmetry for better alignment.}
  \vspace{-0.4cm}
  \label{fig:workflow}
\end{figure}

\section{Introduction}
Composed Image Retrieval (CIR) represents a significant advancement in multimodal search~\cite{Liang_2025,zhang2024gme,kim2025genius,zheng2025onevision}. Unlike single-modality approaches, CIR integrates both reference images and manipulation text as inputs, enabling users to express search intents with enhanced precision. This multimodal paradigm has practical applications, such as e-commerce search, where users express nuanced preferences beyond a single modality.
\begin{figure}[t] 
  \centering 
  \includegraphics[width=\columnwidth]{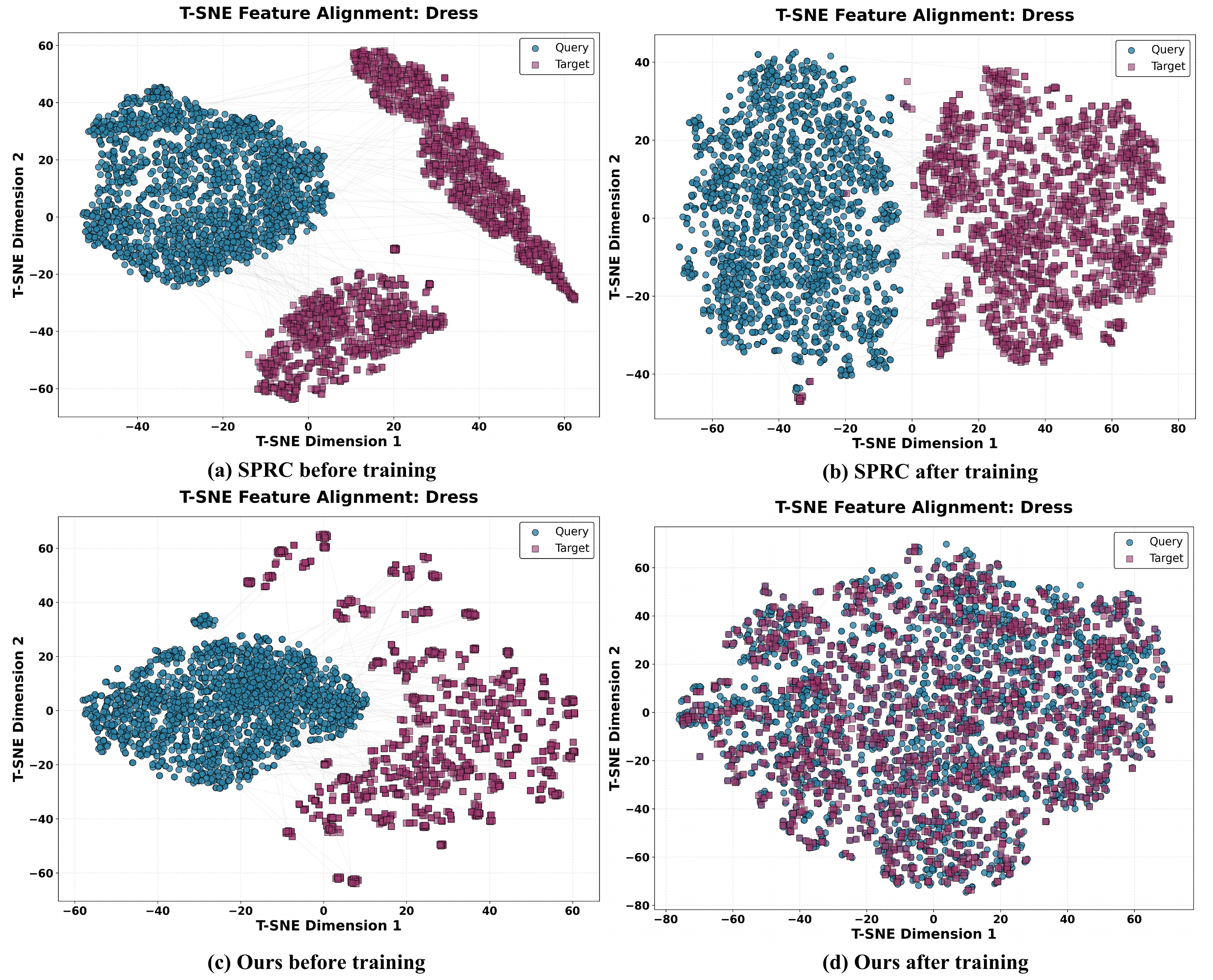}
  \vspace{-0.2cm}
  \caption{T-SNE visualization comparing SPRC and CSMCIR on Fashion-IQ dresses.}
  \vspace{-0.7cm}
  \label{fig:tsne}
\end{figure}

\par
Despite its promising prospects, existing CIR methods suffer from \textbf{representation space fragmentation}~\citep{sun2024leveraging,liu2023candidate,yang2024decomposing,suo2024knowledge}: queries (image and text) and targets (only image) comprise heterogeneous modalities and are processed by distinct encoders, forcing models to bridge misaligned representation spaces only through post-hoc alignment. Researchers have explored various fusion strategies for multimodal integration, including early fusion~\citep{liu2023candidate,levy2024data}, late fusion~\citep{anwaar2021compositional,baldrati2023composed,wen2023target,wen2024simple}, and textual inversion~\cite{baldrati2022effective,saito2023pic2word,baldrati2023zero} approaches. However, as illustrated in Fig.\ref{fig:workflow}(a), these methods maintain asymmetric encoders for query and target sides, perpetuating the fragmentation that fundamentally limits contrastive learning effectiveness. Critically, as shown in Fig.~\ref{fig:tsne}(b), this fragmentation persists even after training, revealing the fundamental limitation of these asymmetric post-hoc alignment strategies. 


\par

\par
Recent advances in Multimodal Large Language Models (MLLMs) enable transforming target images into textual representations through caption generation, addressing the modal inconsistency between query and target sides. However, there still exists a critical challenge: generated captions must be both discriminative to capture distinctive visual attributes and maintain comparable semantic detail with manipulation text. To address this, we propose Multi-level Chain-of-Thought (MCoT) prompting that guides MLLMs through structured reasoning to synthesize captions satisfying these requirements. Unlike approaches~\cite{tian2025ccin,wen2024simple,tang2024reason,lin2025cir} that optimize manipulation prompts at query time and introduce inference latency, our method pre-generates target captions offline, making it practical for real-world deployment while establishing the foundation for modal symmetry.

\par

With target captions generated via MCoT, both query and target sides exhibit consistent multimodal structure as image-text pairs, enabling a symmetric dual-tower architecture with shared-parameter Q-Former (shown in Fig.~\ref{fig:workflow}(b)). By processing both sides through the identical encoder, this design directly addresses the encoder asymmetry in prior methods~\cite{xu2024sentence,tian2025ccin,li2025learning}. The advantage of this symmetric design is evident in Fig.~\ref{fig:tsne}(c): even before training, our method demonstrates a smaller query-target distance, contrasting sharply with the severe fragmentation in the baseline approach. This validates that modal symmetry through consistent encoding establishes a solid foundation for progressive alignment, rather than relying on post-hoc bridging of misaligned spaces.

\par
Finally, the achieved structural symmetry further enables Memory Bank\cite{wu2018unsupervised} integration for enhanced contrastive learning. While prior work~\cite{feng2024improving} deemed Memory Banks unsuitable for CIR due to architectural and modal asymmetry, our symmetric design naturally accommodates this mechanism. However, standard Memory Banks suffer from representation inconsistency in the CIR task. Due to limited training data and the need for rapid parameter updates, the stored representations in Memory Bank become misaligned with current batch representations as the model states evolve. To address this challenge, we propose an entropy-based Memory Bank strategy that incorporates temporal awareness and information-theoretic sample selection. The strategy dynamically updates stored representations using the current model state, ensuring diverse and informative negatives that enhance contrastive learning, improve query-target alignment, and boost training efficiency. As depicted in Fig.~\ref{fig:tsne}(d), after training, our method achieves tightly integrated query-target fusion, with embeddings becoming almost indistinguishable throughout the representation space, proving its effectiveness. 

\par
Taking the above designs into account, we introduce \textbf{CSMCIR}, a \textbf{C}oT-Enhanced \textbf{S}ymmetric Alignment with \textbf{M}emory Bank for \textbf{C}omposed \textbf{I}mage \textbf{R}etrieval. Extensive experiments conducted across Fashion-IQ, CIRR, Shoes and LaSCO datasets demonstrate the effectiveness of our approach. In summary, our contributions include:
\begin{itemize}
\item We propose a unified symmetric framework that systematically addresses representation space fragmentation in CIR through MCoT-enhanced caption generation and parameter-shared dual-tower architecture.

\item We introduce an entropy-based, temporally-aware Memory Bank that maintains representation consistency with evolving model states for enhanced contrastive learning.

\item Extensive experiments across four benchmarks demonstrate that our CSMCIR achieves state-of-the-art performance, with comprehensive ablation studies confirming the effectiveness of each component.
\end{itemize}

\section{Related Work}

\subsection{Composed Image Retrieval}
In recent years, there has been a surge of research and development at the intersection of multimodal learning\cite{zhang2024fast,zhang2025storyweaver,yang2024exploring,qian2024x,qian2024multi} and search technologies\cite{chen2026onesearch}.
CIR enables the retrieval of target images matching both a reference image and manipulation text, requiring an effective understanding of complex semantic interactions between visual and textual modalities.
Early approaches explored various strategies for multimodal fusion, including early-fusion~\cite{liu2023candidate,levy2024data} and late-fusion~\cite{anwaar2021compositional,baldrati2023composed,wen2023target,wen2024simple,chen2025offset,li2025encoder}. Another line of work introduced textual inversion modules~\cite{gal2022image,baldrati2022effective,saito2023pic2word,baldrati2023zero} to transform reference images into pseudo-word embeddings, which are subsequently concatenated with manipulation text for target retrieval.
However, existing methods suffer from representation space fragmentation due to asymmetric query and target modalities. The query side combines a reference image with manipulation text, while the target side contains only an image. This fundamental asymmetry necessitates distinct encoders for each side~\cite{jiang2024cala,wen2023self,xu2024sentence,levy2024data,liu2023candidate,jang2024visual,xing2025context}, forcing models to bridge misaligned representation spaces only through post-hoc alignment, which fundamentally limits contrastive learning effectiveness.
To address this limitation, we propose a unified symmetric dual-tower architecture that establishes consistent representations, directly tackling the representation space fragmentation challenge.

\subsection{Vision and Language Pre-training Models}
Large Vision-Language Models (LVLMs)~\cite{radford2021learning,li2022blip,lu2019vilbert,li2023blip} such as CLIP~\cite{radford2021learning} and BLIP~\cite{li2022blip} have become foundational tools for CIR by enabling alignment between reference images and manipulation  text. Many works~\cite{jiang2024cala,xu2024sentence,wen2023target,wen2023self,ventura2024covr,lin2025cir} leverage LVLMs as encoders to enhance cross-modal matching.
Recent research explores integrating Multimodal Large Language Models (MLLMs)~\cite{li2023blip,liu2023visual,Bai_Bai_Yang_Wang_Tan_Wang_Lin_Zhou_Zhou_2023} into CIR tasks. SPN~\cite{feng2024improving} uses MLLMs to construct training triplets, while CIR-LVLM~\cite{sun2024leveraging} employs them as user intent-aware encoders. Several approaches~\cite{wen2024simple,tang2024reason,sun2023training,karthik2023vision} apply MLLMs to refine manipulation  text at query time for improved matching. However, such query-time text refinement introduces significant inference latency, degrading user experience. In contrast, generating descriptive captions for target images, which can be performed offline, remains largely unexplored.
We argue that MLLMs can produce detailed target image descriptions, enhancing alignment between query and target modalities. To this end, we leverage Chain-of-Thought (CoT) prompting~\cite{wei2022chain,he2024multi,zhang2023multimodal,zheng2023ddcot} to generate enhanced target image captions, establishing unified representation spaces for effective alignment.

\section{Methodology}
\begin{figure*}[t] 
  \centering
\includegraphics[width=2\columnwidth]{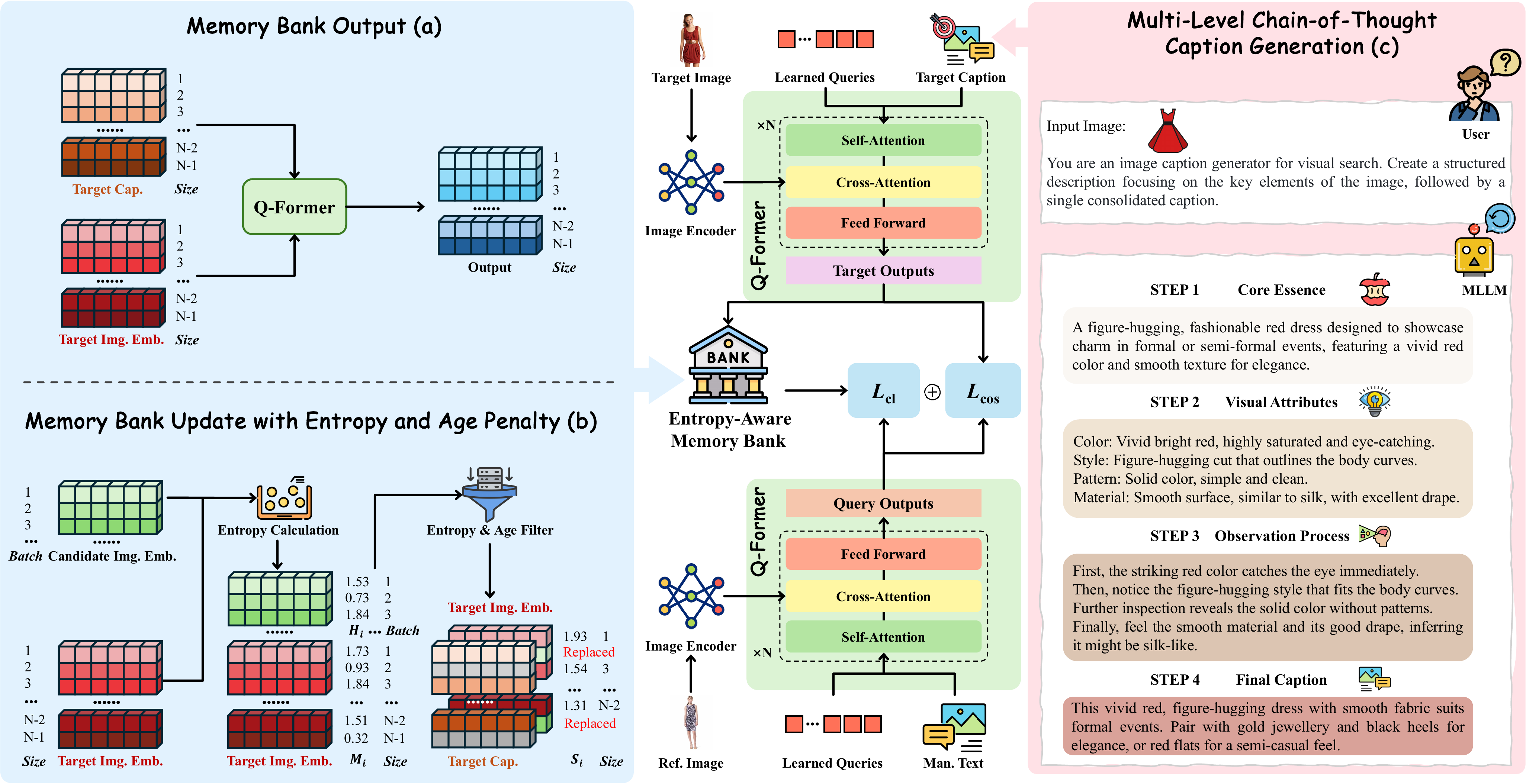}
  \caption{Overview of CSMCIR framework. \textbf{Right:} MCoT-based target caption generation. \textbf{Left:} (a) Entropy-Aware Memory Bank for negative sampling. (b) Memory Bank update via entropy-based and temporal scoring.}
  \vspace{-0.6cm}
  \label{fig:method}
\end{figure*}

\subsection{Preliminary}
CIR addresses the retrieval problem where a query $Q$ combines a reference image $I_r$ with a manipulation text $T$ to search for the relevant target image $I_t$ in a candidate set. Our approach tackles the fundamental limitation of \textbf{representation space fragmentation} in CIR: queries and targets comprise heterogeneous modalities processed by distinct encoders, creating misaligned representation spaces. By employing MLLMs to generate target image captions $T(I_t)$, we transform the paradigm from $(I_r,T,I_t)$ to $(I_r,T,I_t,T(I_t))$ to establish modal symmetry. Notably, this transformation does not alter the essential task definition, as captions are derived solely from target images and generated automatically without any manual annotations. In practical scenario such as e-commerce platforms, these captions can be seamlessly substituted with existing product titles or descriptions.

\subsection{Multi-level Chain-of-Thought Prompting}\label{sec:3.2}
Despite the ease of generating captions for images using MLLMs, caption quality critically impacts model performance. In CIR, users identify target images by describing distinctive differences between reference and target images. Consequently, captions must balance discriminative detail with conciseness: overly brief captions risk omitting critical visual attributes for identification, while exhaustive descriptions introduce redundancy and counterproductive noise that interferes with retrieval. Moreover, semantic and structural inconsistency between manipulation text and target captions may introduce misleading signals that undermine performance.

\par
To address these challenges, we propose a Multi-level Chain-of-Thought (MCoT) prompting strategy that guides MLLMs to generate comprehensive yet discriminative captions with appropriate detail levels for target images. As depicted in Fig.\ref{fig:method}, our MCoT comprises four key steps as follows:

\textbf{Step 1: Core Essence:} The MLLM first generates concise descriptions capturing the essence of target images $I_t$, focusing on main objects and distinctive characteristics.

\textbf{Step 2: Visual Attributes:} Next, the MLLM identifies and describes key visual attributes of objects in the image, including color, material, shape, and spatial relationships.

\textbf{Step 3: Observation Process:} The MLLM then explicates its reasoning process for identifying the Core Essence, detailing which Visual Attributes were prioritized and why.

\textbf{Step 4: Final Caption Formation:} By synthesising insights from the previous steps and a few prompt examples, the MLLM generates a comprehensive final caption that incorporates both primary objects and discriminative details while avoiding redundant descriptions. 

\par
In summary, the complete caption generation process for target images can be formulated as:
\begin{equation}
    T(I_t) = \Psi_{MCoT}(I_t) \quad,
\end{equation}
where $T(I_t)$ denotes the generated caption for image $I_t$, and $\Psi_{MCoT}$ represents the MCoT-based caption generation function.

Our MCoT enhances cross-modal alignment and mitigates representation space fragmentation between query and target sides, establishing a foundation for the symmetric dual-tower architecture and Memory Bank optimization in subsequent training stages. Caption construction requires only a few hours at most and, as a one-time preprocessing step, can be reused across all subsequent experiments.

\subsection{Symmetrical Dual-tower Model Architecture}
After obtaining target image captions, we establish structurally consistent multimodal pairs on both sides: $(I_r, T)$ and $(I_t, T(I_t))$. This modal symmetry enables identical processing operations for both pairs. Specifically, inspired by \cite{xu2024sentence,jiang2024cala}, we adopt the lightweight Querying Transformer (Q-Former) from BLIP-2 \cite{li2023blip} as our cross-modal encoder. As depicted in Fig.\ref{fig:method}, BLIP-2's pretrained image encoder extracts visual features from both reference and target images. These image features, along with their corresponding text (manipulation text for the reference image and generated caption for the target image), are fed into the identical Q-Former with fully shared parameters.

\par
Additionally, a set of learnable query tokens $q$ are introduced to facilitate cross-modal interaction between visual and textual representations on both sides. The encoding process can be formulated as:
\begin{eqnarray}
    Z_q &=& \text{Q-Former}(I_r, T, q),\\
    Z_t &=& \text{Q-Former}(I_t, T(I_t), q),
\end{eqnarray}
where $Z_q$ and $Z_t$ denote the encoded cross-modal representations from the query and target encoders respectively.

\subsection{Entropy-Aware Memory Bank Strategy}
Traditional Memory Bank approaches\cite{wu2018unsupervised} enable efficient large-scale contrastive learning but suffer from temporal inconsistency: frozen representations become misaligned as models evolve, degrading negative sample quality. This limitation is particularly severe in CIR, where representation space fragmentation led prior work (e.g., SPN\cite{feng2024improving}) to deem Memory Banks incompatible. While our symmetric architecture addresses structural misalignment, temporal inconsistency remains a critical challenge.

\noindent\textbf{Static-Dynamic Decoupling.} To address this, we propose a static-dynamic decoupling strategy: the memory bank stores static inputs (captions and frozen image embeddings before being put into Q-Former), while embeddings for negative sampling are dynamically recomputed using the current Q-Former at each step. This ensures all representations remain consistent with the current model state.

\noindent\textbf{Entropy-Aware Sample Selection.} 
Let $\mathcal{B} = \{\mathbf{z}_1, ..., \mathbf{z}_B\}$ denote the [CLS] token embeddings of images from the current batch obtained via ViT (where $B$ is the batch size), and $\mathcal{M} = \{\mathbf{m}_1, ..., \mathbf{m}_M\}$ denote the image embeddings stored in the memory bank (where $M$ is the memory bank size). To enhance sample diversity and select informative hard negatives, we measure uncertainty via information entropy.

Specifically, we first compute similarity-based probability distributions. For batch sample $i$ relative to all memory samples:
\begin{equation}
    p_{i,j}^{\mathcal{B} \rightarrow \mathcal{M}} = \frac{\exp(\mathbf{z}_i^T \mathbf{m}_j)}{\sum_{k=1}^{M} \exp(\mathbf{z}_i^T \mathbf{m}_k)},
\end{equation}
and for memory sample $i$ relative to samples in memory bank:
\begin{equation}
    p_{i,j}^{\mathcal{M} \rightarrow \mathcal{M}} = \frac{\exp(\mathbf{m}_i^T \mathbf{m}_j)}{\sum_{k=1}^{M} \exp(\mathbf{m}_i^T \mathbf{m}_k)}.
\end{equation}

Then we calculate the information entropy to measure the uncertainty of each sample's similarity distribution:
\begin{eqnarray}
    H_i^\mathcal{B} &=& -\sum_{j=1}^{M} p_{i,j}^{\mathcal{B} \rightarrow \mathcal{M}} \log p_{i,j}^{\mathcal{B} \rightarrow \mathcal{M}},\\
    H_i^\mathcal{M} &=& -\sum_{j=1}^{M} p_{i,j}^{\mathcal{M} \rightarrow \mathcal{M}} \log p_{i,j}^{\mathcal{M} \rightarrow \mathcal{M}}.
\end{eqnarray}

Higher entropy indicates that the sample exhibits greater dissimilarity to memory bank samples, making it suitable as an informative negative.

\noindent\textbf{Temporal Decay and Replacement Strategy.} 
To prevent outdated representations from persistently occupying the memory bank, we define a retention score for each memory sample that jointly considers diversity and temporal freshness:
\begin{equation} 
    \hat{H}_i^\mathcal{M} = \underbrace{\max\left(0, 1- \frac{\Delta t_i}{N_{\text{max}}}\right)}_{\text{freshness factor}} \cdot \underbrace{H_i^\mathcal{M}}_{\text{diversity factor}}, 
\end{equation} 
where $\Delta t_i$ denotes the number of training steps since sample $i$ was last updated, and $N_{\text{max}}=10$ controls the maximum staleness threshold before complete decay. This retention score ensures that both low-diversity and stale samples are prioritized for replacement. Finally, batch samples are inserted into the memory bank by replacing memory samples with lower retention entropy, i.e., we replace sample i in the memory bank with sample j in the batch when $H_j^\mathcal{B}> \hat{H}_i^\mathcal{M} $.

Overall, this entropy-aware replacement strategy maintains a diverse and temporally consistent set of negatives, addressing both representation inconsistency and quality degradation inherent in traditional Memory Bank approaches for CIR.

\begin{table*}[t]
    \renewcommand{\arraystretch}{1.3}
    \setlength{\tabcolsep}{9pt}

    \label{tab:1}
    \fontsize{8}{8}\selectfont
    \centering
    \resizebox{\linewidth}{!}{\begin{tabular}{c | cc | cc | cc | ccc}
        \toprule
        
        \multicolumn{1}{c|}{}
        & \multicolumn{2}{c|}{\textbf{Dress}} 
        & \multicolumn{2}{c|}{\textbf{Shirt}} 
        & \multicolumn{2}{c|}{\textbf{Toptee}} 
        & \multicolumn{3}{c}{\textbf{Average}} 
        \\ 
        \cmidrule(lr){2-3} \cmidrule(lr){4-5} \cmidrule(lr){6-7} \cmidrule(lr){8-10}
        
        \multirow{2}{*}[4ex]{\textbf{Method}} 
        & R@10 & R@50 & R@10 & R@50 & R@10 & R@50 & R@10 & R@50 & {Avg.} 
        \\ 
        \cmidrule(lr){1-1}
        \cmidrule(lr){2-3} \cmidrule(lr){4-5} \cmidrule(lr){6-7} \cmidrule(lr){8-10}

        
        



        CoPE~\citep{tang2025modeling} & 39.85 & 66.98 & 45.03 & 66.81 & 48.61 & 72.01 & 44.50 & 68.60 & 56.55 \\
        
        CaLa~\citep{jiang2024cala} & 42.38 & 66.08 & 46.76 & 68.18 & 50.93 & 73.42 & 46.69 & 69.22 & 57.96 \\

        CoVR-BLIP~\citep{ventura2024covr} & 44.55 & 69.03 & 48.43 & 67.42 & 52.60 & 74.31 & 48.53 & 70.25 & 59.39 \\
        
        CASE~\citep{levy2024data} & 47.44 & 69.36 & 48.48 & 70.23 & 50.18 & 72.24 & 48.79 & 70.68 & 59.74 \\
     
        Re-ranking~\citep{liu2023candidate} & {48.14} & {71.43} & {50.15} & {71.25} & {55.23} & {76.80} & {51.17} & {73.13} & {62.15} \\
        
       FashionERN~\citep{chen2024fashionern} & 43.93 & 68.77 & 52.70 & \underline{75.07} & 56.09 & 78.38 & 50.91 & 74.07 & 62.49 \\
       
        SPRC~\citep{xu2024sentence} & 49.18 & 72.43 & 55.64 & 73.89 & \underline{59.35} & 78.58 & 54.72 & 74.97 & 64.85 \\
        
       CCIN~\citep{tian2025ccin} & 49.38 & \underline{72.58} & 55.93 & 74.14 & 57.93 & 77.56 & 54.41 & 74.76 & 64.59 \\

       TME~\citep{li2025learning} & \underline{49.73} & 71.69 & \underline{56.43} & 74.44 & 59.31 & \underline{78.94} & \underline{55.15} & \underline{75.02} & \underline{65.09} \\
       
        \cmidrule(lr){1-10} 
       \rowcolor{gray!25}{ \textbf{CSMCIR(Ours)}} 
        & {\textbf{52.45}} & {\textbf{74.81}} 
        & {\textbf{57.70}} & {\textbf{75.76}} 
        & {\textbf{61.14}} & {\textbf{80.98}} 
        & {\textbf{57.07}} & {\textbf{77.27}} 
        & {\textbf{67.17}} \\
        \bottomrule

    \end{tabular}}
    \vspace{-8pt}
    \caption{\small \textbf{Quantitative} comparison on the \textbf{Fashion-IQ} \texttt{validation} set. Overall $1^{st}$ /$2^{nd}$ in bold/underline.}
    \vspace{-8pt}
    \label{tab:fashioniq}
\end{table*}
\begin{table*}[t]
\renewcommand{\arraystretch}{1.3}
\setlength{\tabcolsep}{9pt}
\label{tab:2}
\fontsize{8}{8}\selectfont
\centering
\resizebox{\linewidth}{!}{\begin{tabular}{c |cccc| ccc |c } 
\toprule

\multicolumn{1}{c|}{}
& \multicolumn{4}{c|}{\textbf{Recall@K}} 
& \multicolumn{3}{c|}{\textbf{Recall{\tiny subset}@K}} 
& \multirow{2}{*}{\textbf{$(R@5+R_{sub}@1)/2$}\textit{}}
\\ 
\cmidrule(lr){2-5} \cmidrule(lr){6-8} 

\multirow{2}{*}[4ex]{\textbf{Method}} 
& K=1 & K=5 & K=10 & K=50 
& K=1 & K=2 & K=3 
&  \\
\cmidrule(lr){1-1}
\cmidrule(lr){2-2} \cmidrule(lr){3-3} \cmidrule(lr){4-4} \cmidrule(lr){5-5}
\cmidrule(lr){6-6} \cmidrule(lr){7-7} \cmidrule(lr){8-8}
\cmidrule(lr){9-9}




CoPE~\citep{tang2025modeling} & 49.18 & 80.65 & 89.86 & 98.05 & 72.34 & 88.65 & 95.30 & 76.49 \\
CASE~\citep{levy2024data} & 48.00 & 79.11 & 87.25 & 97.57 & 75.88 & 90.58 & 96.00 & 77.50 \\
CaLa~\citep{jiang2024cala} & 49.11 & 81.21 & 89.59 & 98.00 & 76.27 & 91.04 & 96.46 & 78.74 \\
CoVR-BLIP~\citep{ventura2024covr} & 49.69 & 78.60 & 86.77 & 94.31 & 75.01 & 88.12 & 93.16 & 80.81 \\
Re-ranking~\citep{liu2023candidate} & {50.55} & {81.75} & {89.78} & {97.18} & {80.04} & {91.90} & {96.58} & {80.90} \\
{{SPRC~\citep{xu2024sentence}}} 
& {51.96} & {82.12} & {89.74} & {97.69} 
& {80.65} & {92.31} & {96.60} 
& {81.39} \\

CCIN~\citep{tian2025ccin} & {53.41} & \underline{84.05} & \underline{91.17} & {98.00} & {-} & {-} & {-} & {-} \\
TME~\citep{li2025learning} & \underline{53.42} & {82.99} & {90.24} & \underline{98.15} & \textbf{81.04} & \underline{92.58} & \underline{96.94} & \underline{82.01} \\

\cmidrule(lr){1-9} 
\rowcolor{gray!25}[\tabcolsep][\tabcolsep] {\textbf{CSMCIR(Ours)}} 
& \textbf{53.88} & \textbf{85.06} & \textbf{91.69} & \textbf{98.58} 
& \underline{80.10} & \textbf{92.24} & \textbf{96.92} 
& \textbf{82.58} \\
\bottomrule
\end{tabular}}
\vspace{-8pt}
\caption{\small \textbf{Quantitative} comparison on the \textbf{CIRR} \texttt{test} set. Overall $1^{st}$ /$2^{nd}$ in bold/underline.}
\vspace{-12pt}
\label{tab:cirr}
\end{table*}
\begin{table}[ht]
    \centering
    \resizebox{\columnwidth}{!}{%
        \begin{tabular}{lcccc}
            \hline
            \textbf{Method} & R@1 & R@10 & R@50 & Avg. \\ \hline
            FashionERN~\citep{chen2024fashionern} & - & 55.59 & 81.71 & - \\
            Prog. Lrn.~\citep{zhao2022progressive} & 22.88&58.83 &84.16 &55.29 \\
            CAFF~\citep{wan2024cross} & 25.21 & 60.17 & 80.79 & 55.39 \\
            TG-CIR~\citep{wen2023target}  & 25.89 & 63.20 & 85.07 & 58.05 \\
            CCIN~\citep{tian2025ccin} & 25.95 & 65.76 & 86.54 & 59.42 \\
            \rowcolor{gray!25} \textbf{CSMCIR(Ours)}  &\textbf{27.83} & \textbf{68.37} & \textbf{87.90} & \textbf{61.37} \\
            \hline
        \end{tabular}
    }
      \vspace{-8pt}
    \caption{Results on Shoes validation set. The best results are highlighted in bold.}
    \vspace{-8pt}
    \label{tab:shoes_val}

\end{table}
\begin{table}[!t]
\begin{center}
\resizebox{\columnwidth}{!}{%
\begin{tabular}{lccccccc}
\hline
\textbf{Method}& R@1 & R@5 & R@10 & R@50 & R@500 & Avg. \\ \hline
Random  & 0.00 & 0.01 & 0.03 & 0.13 & 1.26 &0.72 \\
LF-CLIP~\citep{baldrati2022effective} &4.01 & 10.23 & 14.68 & 32.08 & 72.69 & 26.74 \\
LF-BLIP~\citep{baldrati2022effective}   & 4.26 & 12.01 & 17.11 & 36.54 & 74.62 & 28.91 \\
CASE~\citep{levy2024data} & {7.08} & {18.50} & {26.16} & {50.25} & {85.46} & {37.49} \\
\rowcolor{gray!25} \textbf{CSMCIR(Ours)}  &\textbf{7.83} & \textbf{23.89} & \textbf{32.90} & \textbf{58.44} & \textbf{90.31} & \textbf{42.68} \\
 \hline
\end{tabular}

}
\vspace{-8pt}
\caption{Results on LaSCO validation set. The best results are highlighted in bold. }
\vspace{-12pt}
\label{tab:lasco_val}
\end{center}
\end{table}

\subsection{Learning Objectives}
Following previous works, contrastive loss is introduced to achieve alignment between the query and target sides of the CIR task. Specifically, following \cite{xu2024sentence}, we utilize the [CLS] token $e_{cls}$ from the query side's output $Z_q$ as our query embedding $u$, which encapsulates global information of the query encoder output. For the target embedding $v$, we select the query token from the target side's output $Z_t$ with the highest similarity to the query embedding $u$.

\subsubsection{Memory Bank Enhanced Contrastive Loss}
For query-to-target alignment, since we implement the Memory Bank approach, the Memory Bank Enhanced contrastive loss is defined as follows:
\begin{equation}
\mathcal{L}_{cl}=-\frac{1}{|{B}|} \sum_{i \in {B}} \log \frac{\exp \left(\tau {u}_i^T {v}_i\right)}{\sum_{j \in {B^*}} \exp \left(\tau {u}_i^T {v}_j\right)},
\end{equation}
where ${B}$ denotes the standard batch and ${B^*}$ represents the batch expanded by our Entropy-Aware Memory Bank.

\subsubsection{Adaptive Cosine Loss Alignment}
Since on the target side, the above-mentioned contrastive loss only selects the query token with the highest similarity with the query embedding $u$ as our target embedding $v$, this loss focuses solely on the most relevant token while potentially overlooking valuable information contained in other query tokens on the target side.
To better leverage the multimodal information from the target side, we further introduce alignment between the query embedding $u$ and all query tokens on the target side. Specifically, we introduce a learnable tensor $\alpha$ (initialized to 1) to adaptively weight the importance of different query tokens on the target side, then apply average pooling to aggregate the query tokens, and finally adopt a cosine loss to ensure comprehensive alignment between the two sides:

\begin{equation}
\mathcal{L}_{ {cos}}=\frac{1}{B} \sum_{i=1}^{B}\left(1-\frac{\left(\frac{1}{K} \sum_{k=1}^{K} \alpha_{k} \cdot v_{k}^{i}\right) \cdot u^{i}}{\left\|\frac{1}{K} \sum_{k=1}^{K} \alpha_{k} \cdot v_{k}^{i}\right\| \cdot\left\|u^{i}\right\|}\right),
\end{equation}
where $K$ represents the number of query tokens.

Finally, the overall loss $\mathcal{L}$ is formulated as:
\begin{equation}
\mathcal{L}=\mathcal{L}_{cl}+\mathcal{L}_{cos}.
\end{equation}

\begin{figure*}[!ht] 
  \centering 
  \includegraphics[width=2\columnwidth]{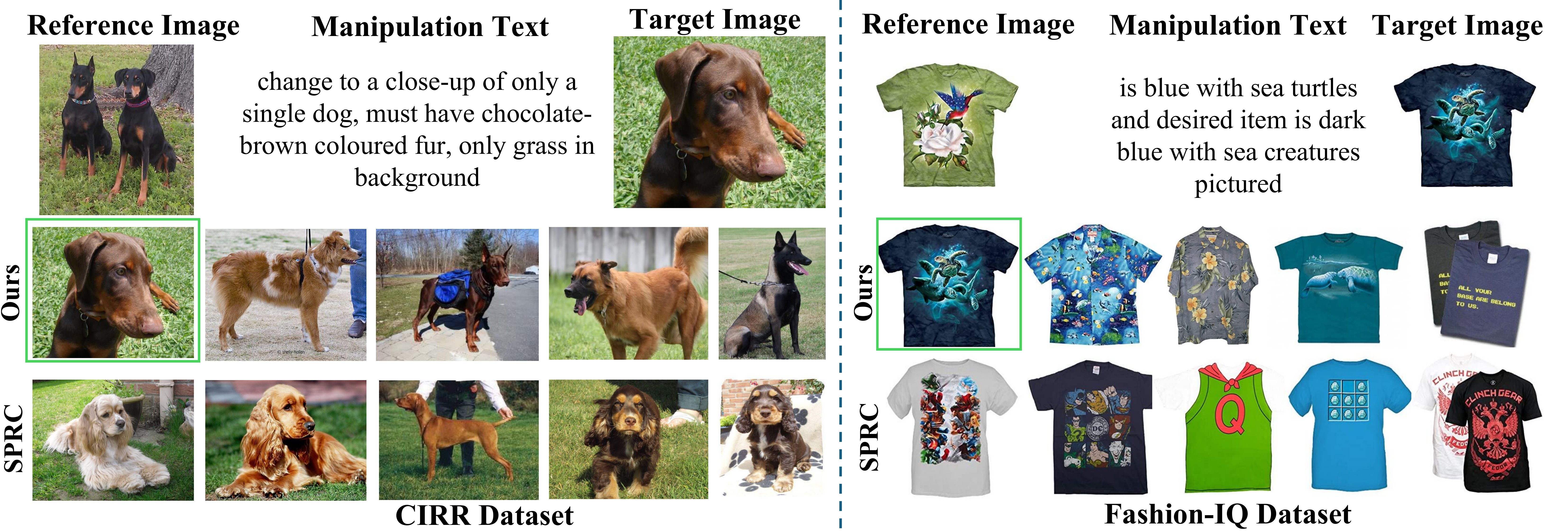}
  \vspace{-0.3cm}
  \caption{Qualitative comparison with SPRC. Green outlines indicate successfully retrieved targets.}
  \vspace{-0.5cm}
  \label{fig:qt}
\end{figure*}

\section{Datasets and Evaluation Metrics}
We evaluate our approach on three standard CIR benchmarks: Fashion-IQ, CIRR, Shoes and LaSCO. Fashion-IQ contains 77,684 fashion images forming 30,134 triplets across three categories (Dress, Toptee, and Shirt), with performance measured using Recall@10, Recall@50 and Recall$_{mean}$ on the validation set. CIRR offers a more diverse benchmark with 36,554 triplets from 21,552 natural images, featuring everyday object interactions. CIRR evaluation uses Recall@1,5,10,50 for general performance and includes a specialized Recall$_{subset}$@1,2,3 metric for a challenging subset containing visually similar distractors, testing fine-grained discrimination capabilities. 
Shoes dataset is divided into 10K triplets for training and 4.6K for testing, performance measured using Recall@1, Recall@10, Recall@50 and Recall$_{mean}$.
LaSCO is a dataset based on COCO images and VQA2.0 annotations, containing 389,305 queries on 121,479 natural images. Compared to CIRR, LaSCO offers x10 more queries, x2 more unique tokens, and x17 more corpus images across an open and broad domain of natural images with rich text. Performance on LaSCO is evaluated using Recall@1,5,10,50,500 and Recall$_{mean}$ metrics.
For LaSCO dataset, we use its own image captions as target image captions.
\section{Experiments}
\subsection{Implementation Details}
Our framework was implemented in PyTorch. Following \cite{xu2024sentence}, we adopt ViT-G as the image encoder, which remained frozen throughout training. We utilized the AdamW optimizer\cite{loshchilov2017decoupled} with a weight decay of 0.05 and a learning rate of 1e-5 on a cosine decay schedule. For our model configuration, we set the number of query tokens to 32 and employed a batch size of 128. The model was trained for 15 epochs with Memory Bank capacities of 512 for CIRR, Fashion-IQ and Shoes datasets, and the Memory Bank capacity is 640 for the LaSCO dataset, age threshold $N_{max}$ is  10. For Fashion-IQ, Shoes and CIRR datasets, we utilize Qwen2.5-VL-7B-Instruct \cite{bai2025qwen2} for target image caption generation. For the LaSCO dataset, we directly used the captions from its VQA2.0 annotations as target image descriptions.
\subsection{Comparision Results}
\subsubsection{Quantitative Results}
Tab.~\ref{tab:fashioniq} shows CSMCIR achieves the highest recall across all Fashion-IQ metrics. Compared to TME\cite{li2025learning} (which also uses Q-Former), our method delivers substantial gains (\textbf{65.09} vs. \textbf{67.17} in Avg. metric). As shown in Tab.~\ref{tab:shoes_val}, on the Shoes dataset, CSMCIR significantly outperforms CCIN \cite{tian2025ccin} by \textbf{1.95} points. Tab.~\ref{tab:cirr} and Tab.~\ref{tab:lasco_val} show that CSMCIR also achieves state-of-the-art performance on open-domain CIRR and LaSCO datasets. Despite CIRR's increased complexity, our method excels particularly in the (R@5+R$_{sub}$@1)/2 metric, substantially outperforming TME\cite{li2025learning}
(\textbf{82.58} vs. \textbf{82.01}), demonstrating strong generalizability. On LaSCO, CSMCIR achieves impressive gains of \textbf{6.74} and \textbf{8.19} in R@10 and R@50 metrics respectively.
These consistent improvements across diverse datasets confirm CSMCIR's effectiveness and robustness for composed image retrieval tasks.

Beyond performance metrics, training efficiency is crucial for practical utility. Our streamlined architecture with Memory Bank not only enhances performance but also significantly reduces training costs. Using identical configurations (Q-Former with ViT-G backbone), our method requires only \textbf{1.2} GPU hours on Fashion-IQ compared to SPRC's \textbf{1.8} hours, and \textbf{2.1} GPU hours on CIRR versus SPRC's \textbf{4.6} hours. Our CSMCIR also achieves superior inference efficiency, with a latency of 0.03s per item (faster than SPRC's 0.035s) and memory consumption of 6845MB (lower than SPRC's 7140MB), making it more practical for real-world deployment.

\subsubsection{Qualitative Results}

Fig. \ref{fig:qt} visualizes top-5 retrieval results on CIRR and Fashion-IQ datasets. Our CSMCIR demonstrates superior attention to image details compared to SPRC. On CIRR, we not only correctly predict the target at R@1, but also retrieve dogs with consistent breeds across all top-5 results, while SPRC shows significant breed variations. On Fashion-IQ, our top-1, top-2, and top-4 results closely match the manipulation text description, whereas SPRC fails to do so.

\begin{table}[t]
\begin{center}
\resizebox{\columnwidth}{!}{%
\begin{tabular}{l ccc @{\hspace{8pt}} ccc}
\hline
\multirow{2}{*}{\textbf{Method}} & \multicolumn{3}{c}{\textbf{Fashion-IQ}} & \multicolumn{3}{c}{\textbf{CIRR}} \\
\cmidrule(lr){2-4} \cmidrule(lr){5-7}
& R@10 & R@50 & Avg. & R@5 & 
R$_{sub}$@1& Avg.\\

\hline
baseline & 54.92 & 74.97 & 64.85 &82.12 & 80.65 & 81.39\\
+MCoT & 55.41 & 75.25 & 65.33 &83.61 & 79.68 & 81.65\\
+SA & 55.26 & 76.57 & 65.92 &83.89  & 79.96 & 81.93\\
+EAMB &56.52 &76.78  & 66.65 & 84.15 & 80.14 & 82.15\\
\rowcolor{gray!25} +$L_{\text{cos}}$ & \textbf{57.07} & \textbf{77.27} & \textbf{67.17} & \textbf{85.06} & \textbf{80.10} & \textbf{82.58}\\
\hline
\end{tabular}
}
\vspace{-8pt}
\caption{The results obtained after ablating different
modules on the Fashion-IQ and CIRR datasets. }
\label{tab:ablation}
\end{center}
\vspace{-8pt}
\end{table}

\begin{table}[t]
\begin{center}
\resizebox{\columnwidth}{!}{%
\begin{tabular}{l ccc @{\hspace{8pt}} ccc}
\hline
\multirow{2}{*}{\textbf{Method}} & \multicolumn{3}{c}{\textbf{Fashion-IQ}} & \multicolumn{3}{c}{\textbf{CIRR}} \\
\cmidrule(lr){2-4} \cmidrule(lr){5-7}
& R@10 & R@50 & Avg. & R@5 & R$_{sub}$@1& Avg.\\
\hline
baseline & 54.92 & 74.97 & 64.85 &82.12 & 80.65 & 81.39\\
w/o TD & 55.79 & 76.49 & 66.14 &84.25  & 80.06 & 82.16\\
w/o EA &56.33 &76.95  & 66.64 & 84.35 & 80.16 & 82.26\\
\rowcolor{gray!25} Full & \textbf{57.07} & \textbf{77.27} & \textbf{67.17} & \textbf{85.06} & \textbf{80.10} & \textbf{82.58}\\
\hline
\end{tabular}
}
\vspace{-10pt}
\caption{Ablation studies on EAMB Strategy. }
\vspace{-15pt}

\label{tab:ablation2}
\end{center}
\vspace{-8pt}
\end{table}

\subsection{Ablation Study}
To evaluate the effectiveness of our designed model architecture and modules, we conducted extensive experiments on the test set of CIRR and the validation set of Fashion-IQ.
\subsubsection{Modules Ablation Study}
As shown in Tab.\ref{tab:ablation}, since our model is built on SPRC\cite{xu2024sentence}, we set its performance as our baseline. Specifically, our Multi-level Chain-of-Thought(MCoT) successfully establishes cross-modal alignment by transforming the retrieval paradigm from asymmetric to symmetric, thereby achieving superior query-target alignment. Building upon this foundation, as shown in line3, our Symmetrical Structure (SA) further enhances alignment by employing identical shared-parameter encoders on both sides, delivering remarkable performance improvements. This validates the effectiveness of our unified architecture in addressing the modal alignment challenges inherent in CIR tasks.
Furthermore, we validate our Entropy-Aware Memory Bank strategy (EAMB), which generates high-quality negative samples and expands batch size, achieving an improvement from 65.92 to 66.65 in the Fashion-IQ dataset. When combining Memory Bank Enhanced Contrastive Loss and Adaptive Cosine Loss, our framework further achieves performance gains, confirming their effectiveness.
\begin{figure}[!t] 
  \centering 
  \includegraphics[width=\columnwidth]{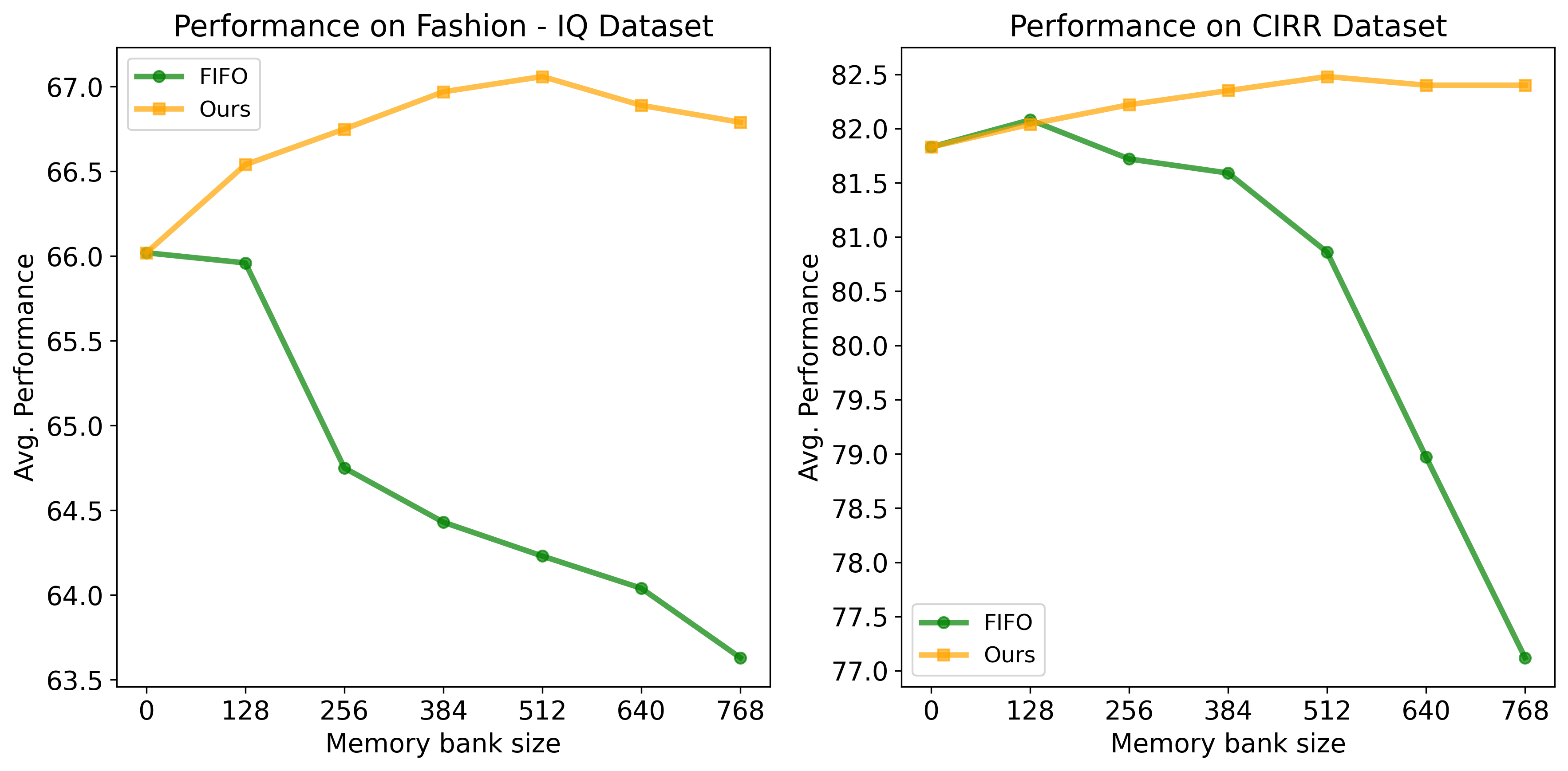} 
  \vspace{-0.65cm}
  \caption{Ablation study on Memory Bank strategies.  }
  \vspace{-0.65cm}
  \label{fig:mb}
\end{figure}

\begin{table*}[ht]
    \centering
    \footnotesize 
    \renewcommand{\arraystretch}{1.2} 
    
    \setlength{\tabcolsep}{1pt}
    
    \resizebox{\linewidth}{!}{\begin{tabular}{@{} l | ccc | ccc | cccc | cccc @{}}
        \toprule
        
        \multirow{2}{*}{\textbf{Method}} 
        & \multicolumn{3}{c|}{\textbf{Fashion-IQ}} 
        & \multicolumn{3}{c|}{\textbf{CIRR}} 
        & \multicolumn{4}{c|}{\textbf{Shoes}} 
        & \multicolumn{4}{c}{\textbf{Lasco}} \\
        
        \cmidrule(lr){2-4} \cmidrule(lr){5-7} \cmidrule(lr){8-11} \cmidrule(lr){12-15}
        
        & R@10 & R@50 & Avg. 
        & R@5 & R$_{\text{sub}}$@1 & Avg. 
        & R@1 & R@10 & R@50 & Avg. 
        & R@5 & R@10 & R@50 & Avg. \\
        \midrule
        
       CoPE (ViT-L)\cite{tang2025modeling}  & 44.50 & 68.60 & 56.55 & 80.65 & 72.34 & 76.49 & {-} & {-} & {-} & {-} & {-} & {-} & {-} & {-} \\        
       SPRC (ViT-L)\cite{xu2024sentence}  & 51.04 & 73.38 & 62.21 & 80.65 & 79.59 & 80.12 & {-} & {-} & {-} & {-} & {-} & {-} & {-} & {-} \\
       QURE (ViT-L)\cite{kwak2025qure}  & 52.60 & 73.48 & 63.04 & 82.53 & 78.51 & 80.52 & {-} & {-} & {-} & {-} & {-} & {-} & {-} & {-} \\
      \rowcolor{gray!25} Ours (ViT-L)  & 54.25 & 74.66 & 64.45 & 83.04 & 78.82 & 80.93 & {25.50} & {64.74} & {87.11} & {59.11} & {22.45} & {31.35} & {56.93} & {36.91} \\
        \midrule
       \rowcolor{lightgray} \textbf{Ours (ViT-G)}  & \textbf{57.07} & \textbf{77.27} & \textbf{67.17} & \textbf{85.06} & \textbf{80.10} & \textbf{82.58} & \textbf{27.83} & \textbf{68.37} & \textbf{87.90} & \textbf{61.37} & \textbf{23.89} &\textbf{32.90} & \textbf{58.44} & \textbf{38.14} \\
        \bottomrule
    \end{tabular}}
   
    \caption{ViT type ablations. Our method demonstrates superior performance across different ViT architectures, including ViT-L, proving its robustness and stability. Furthermore, our method achieves significant performance improvements when scaling from ViT-L to ViT-G architectures.}
    \vspace{-10pt}
     \label{tab:vit_ablation}
\end{table*}
\subsubsection{Memory Bank Ablations}
We conducted ablation experiments comparing our Memory Bank strategy with the naive FIFO approach from MoCo~\cite{he2020momentum}. Unlike traditional representation learning, CIR involves fewer samples, more complex modalities, and faster model updates. Directly storing Q-former-generated cross-modal features proves problematic: rapid Q-former updates across batches create significant feature distribution disparities, rendering them unsuitable as negatives for subsequent batches.
As shown in Fig.~\ref{fig:mb}, the FIFO strategy exhibits rapid performance degradation when Memory Bank size exceeds batch size. Our strategy addresses this by storing captions and image embeddings from previous batches, then constructing negatives through Q-Former in the current batch to maintain feature distribution consistency. Performance plateaus at Memory Bank size 512, likely sufficient to capture the feature distributions of the Fashion-IQ and CIRR datasets.
We also conduct ablation studies on the EAMB strategy components, with results shown in Table~\ref{tab:ablation2}. Removing either Temporal Decay (TD) or Entropy-Aware (EA) degrades performance on both Fashion-IQ and CIRR benchmarks, with TD showing greater impact. The full model achieves the best results, confirming the effectiveness of both components.

\subsubsection{ViT Ablation Study}
The image encoder plays a vital role in image feature extraction, significantly impacting overall model performance. As shown in Tab.\ref{tab:vit_ablation}, across all three datasets: Fashion-IQ, CIRR, Shoes and LaSCO, our model's performance with ViT-G substantially outperforms the ViT-L version, confirming the importance of high-quality visual features in the CIR task. Nevertheless, even with the ViT-L version, our model still achieves excellent results, significantly outperforms QURE\cite{kwak2025qure} and CoPE\cite{tang2025modeling} under equivalent conditions, demonstrating the effectiveness of our approach.

\subsubsection{$N_{max}$ parameter Ablation}
\label{B2}
\begin{figure}[t]
  \centering
\includegraphics[width=\columnwidth]{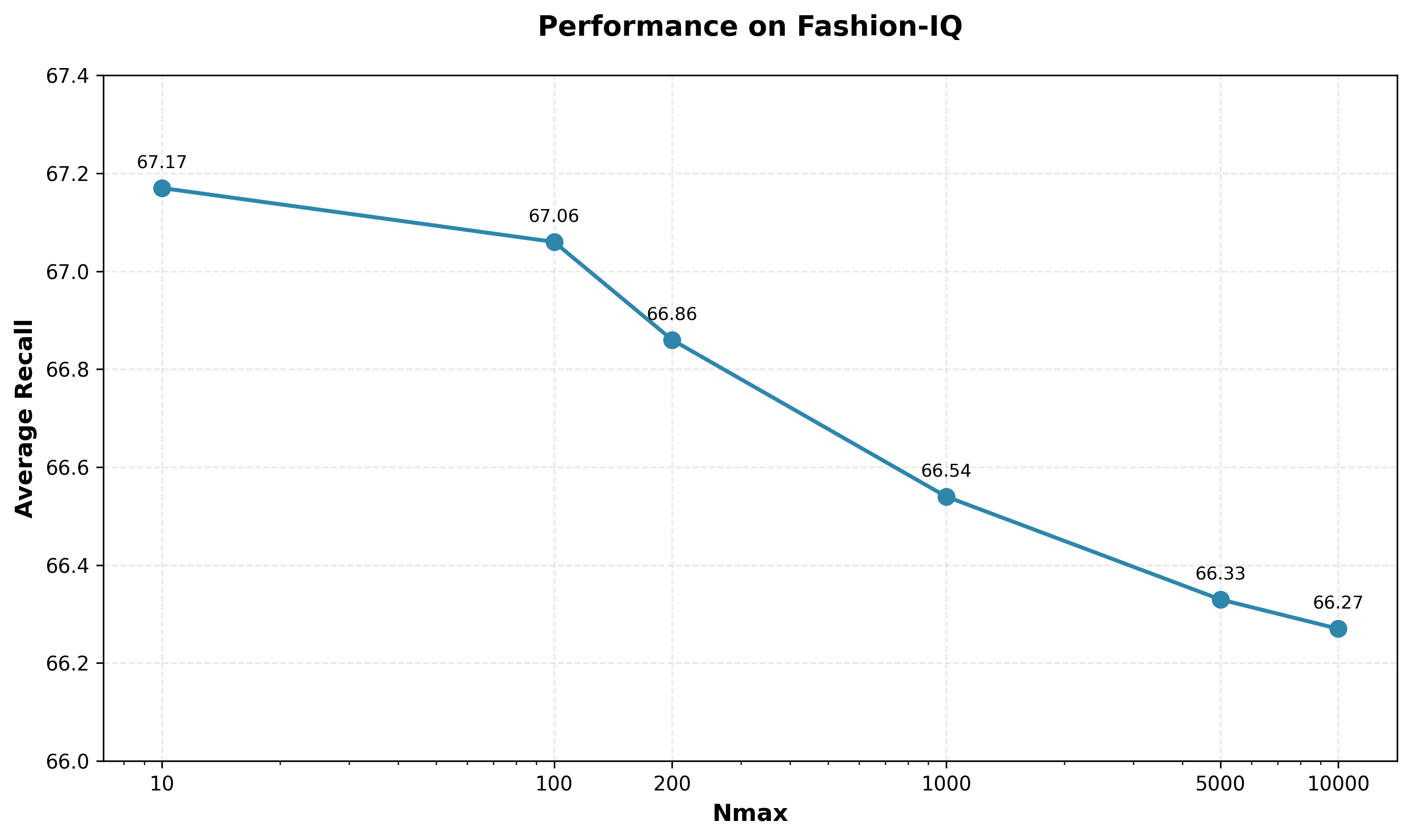}
  \vspace{-0.2cm}
  \caption{Ablation study on $N_{max}$ parameter in our Memory Bank strategy on Fashion-IQ dataset.}
  \vspace{-0.2cm}
  \label{fig:nmax}
\end{figure}

 As shown in Fig.~\ref{fig:nmax}, the performance drops as $N_{max}$ increases, which validates the effectiveness of our time-step-based update strategy. This indicates that although entropy-based selection can identify high-quality negative samples to some extent, relying on these samples without timely updates significantly degrades the effectiveness of the memory bank strategy.

\section{Conclusion}
In this work, we propose CSMCIR, a unified symmetric framework that systematically addresses representation space fragmentation in CIR. Our approach achieves state-of-the-art performance through three synergistic innovations: Multi-level Chain-of-Thought prompting, parameter-shared dual-tower architecture, and Entropy-Aware Memory Bank. Extensive experiments across four benchmarks demonstrate CSMCIR's superior performance, with comprehensive ablation studies confirming the effectiveness of each component. Our code is available at \url{https://github.com/qzp2018/CSMCIR}.

\section{Limitations}
While our CSMCIR framework achieves state-of-the-art performance on multiple benchmarks, several limitations warrant discussion:
\subsection{Caption Generation Dependency:}Our approach leverages MLLMs to pre-generate captions for target images in an offline manner. While this effectively eliminates inference latency during runtime, it entails additional upfront time investment to generate the required captions for datasets where such target image captions are not readily available.
\subsection{Dataset Scope:}Our evaluation is centered on the fashion and general object domains, with experiments conducted on benchmark datasets including Fashion-IQ, CIRR, Shoes, and LaSCO. However, performance in specialized domains—such as medical imaging and fine art—remains uninvestigated, primarily due to the scarcity of task-specific, high-quality datasets.
\newpage
\bibliography{custom}

\clearpage
\appendix

\section{Entropy-Aware Memory Bank Strategy Flowchart}
\label{C}

Our approach employs a \textbf{static-dynamic decoupling mechanism}: the memory bank stores static inputs (image captions $T(I_t)$ generated by MLLMs and frozen image embeddings from ViT), while during training, embeddings for negative sampling are dynamically recomputed using the current Q-Former. This ensures all representations remain consistent with the current model state, resolving the temporal inconsistency problem in traditional memory banks.

Algorithm \ref{alg:memory_update} presents our entropy-aware memory bank update strategy in five key steps: First, it extracts batch embeddings $\mathcal{B}$ via ViT (Step 1). Second, it computes similarity-based probability distributions between batch samples and memory samples, as well as within the memory bank itself (Step 2). Third, information entropy $H_i^\mathcal{B}$ and $H_i^\mathcal{M}$ are calculated to quantify the diversity of each sample—higher entropy indicates greater dissimilarity to existing memory samples, making them informative hard negatives (Step 3). Fourth, a retention score $\hat{H}_i^\mathcal{M}$ is computed by jointly considering both diversity (entropy factor) and temporal freshness (decay factor based on $\Delta t_i$), ensuring stale samples are prioritized for replacement (Step 4). Finally, high-entropy batch samples replace low-scoring memory samples when $H_j^\mathcal{B} > \hat{H}_i^\mathcal{M}$, maintaining a diverse and temporally consistent negative sample pool (Step 5). This comprehensive strategy effectively maintains memory bank quality while addressing representation inconsistency, enabling efficient large-scale contrastive learning for CIR tasks.

\begin{figure}[h]
  \centering
\includegraphics[width=\columnwidth]{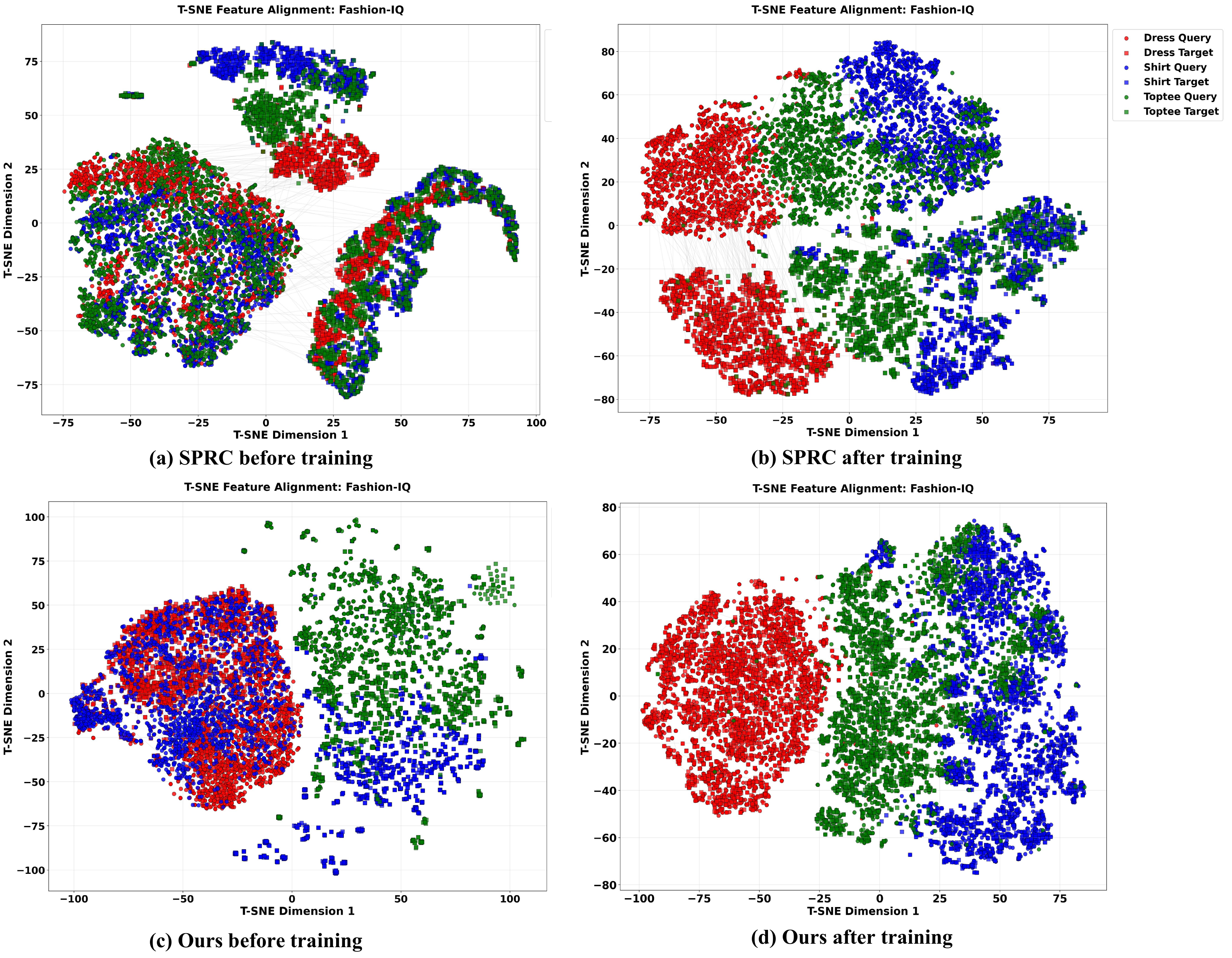}
  \vspace{-0.2cm}
  \caption{T-SNE visualization comparing SPRC and
CSMCIR on the Fashion-IQ's all three categories.}
  \vspace{-0.2cm}
  \label{fig:tsne_2}
\end{figure}

\begin{algorithm}[h]
\caption{Entropy-Aware Memory Bank Update}
\label{alg:memory_update}
\small
\begin{algorithmic}[1]
\STATE \textbf{Input:} Batch images $\{{I}_i\}_{i=1}^B$, image captions $T(I_t)$, memory bank $\mathcal{M}$
\STATE \textbf{Output:} Updated memory bank $\mathcal{M}$

\STATE \textcolor{blue}{// Step 1: Extract batch embeddings via ViT}
\STATE $\mathcal{B} = \{\mathbf{z}_1, ..., \mathbf{z}_B\}$ where $\mathbf{z}_i = \text{ViT}({I}_i)$

\STATE \textcolor{blue}{// Step 2: Compute similarity-based probability distributions}
\FOR{each batch sample $i \in [1, B]$}
    \STATE $p_{i,j}^{\mathcal{B} \rightarrow \mathcal{M}} = \exp(\mathbf{z}_i^T \mathbf{m}_j) / \sum_{k=1}^{M} \exp(\mathbf{z}_i^T \mathbf{m}_k)$, $\forall j \in [1,M]$
\ENDFOR
\FOR{each memory sample $i \in [1, M]$}
    \STATE $p_{i,j}^{\mathcal{M} \rightarrow \mathcal{M}} = \exp(\mathbf{m}_i^T \mathbf{m}_j) / \sum_{k=1}^{M} \exp(\mathbf{m}_i^T \mathbf{m}_k)$, $\forall j \in [1,M]$
\ENDFOR

\STATE \textcolor{blue}{// Step 3: Calculate information entropy for diversity}
\FOR{each batch sample $i \in [1, B]$}
    \STATE $H_i^\mathcal{B} = -\sum_{j=1}^{M} p_{i,j}^{\mathcal{B} \rightarrow \mathcal{M}} \log p_{i,j}^{\mathcal{B} \rightarrow \mathcal{M}}$
\ENDFOR
\FOR{each memory sample $i \in [1, M]$}
    \STATE $H_i^\mathcal{M} = -\sum_{j=1}^{M} p_{i,j}^{\mathcal{M} \rightarrow \mathcal{M}} \log p_{i,j}^{\mathcal{M} \rightarrow \mathcal{M}}$
\ENDFOR

\STATE \textcolor{blue}{// Step 4: Compute retention score with temporal decay}
\FOR{each memory sample $i \in [1, M]$}
    \STATE $\hat{H}_i^\mathcal{M} = \max(0, 1- \Delta t_i / N_{\text{max}}) \cdot H_i^\mathcal{M}$
\ENDFOR

\STATE \textcolor{blue}{// Step 5: Replace low-scoring memory samples}
\STATE Sort batch samples by $H^\mathcal{B}$ (descending)
\STATE Sort memory samples by $\hat{H}^\mathcal{M}$ (ascending)
\FOR{each high-entropy batch sample $j$}
    \IF{$H_j^\mathcal{B} > \hat{H}_i^\mathcal{M}$ for lowest-scoring memory sample $i$}
        \STATE $\mathbf{m}_i \leftarrow \mathbf{z}_j$, $T(I_t)_i \leftarrow T(I_t)_j$, $\Delta t_i \leftarrow 0$
    \ENDIF
\ENDFOR

\RETURN Updated $\mathcal{M}$
\end{algorithmic}
\end{algorithm}

\begin{figure*}[t]
  \centering
\includegraphics[width=2\columnwidth]{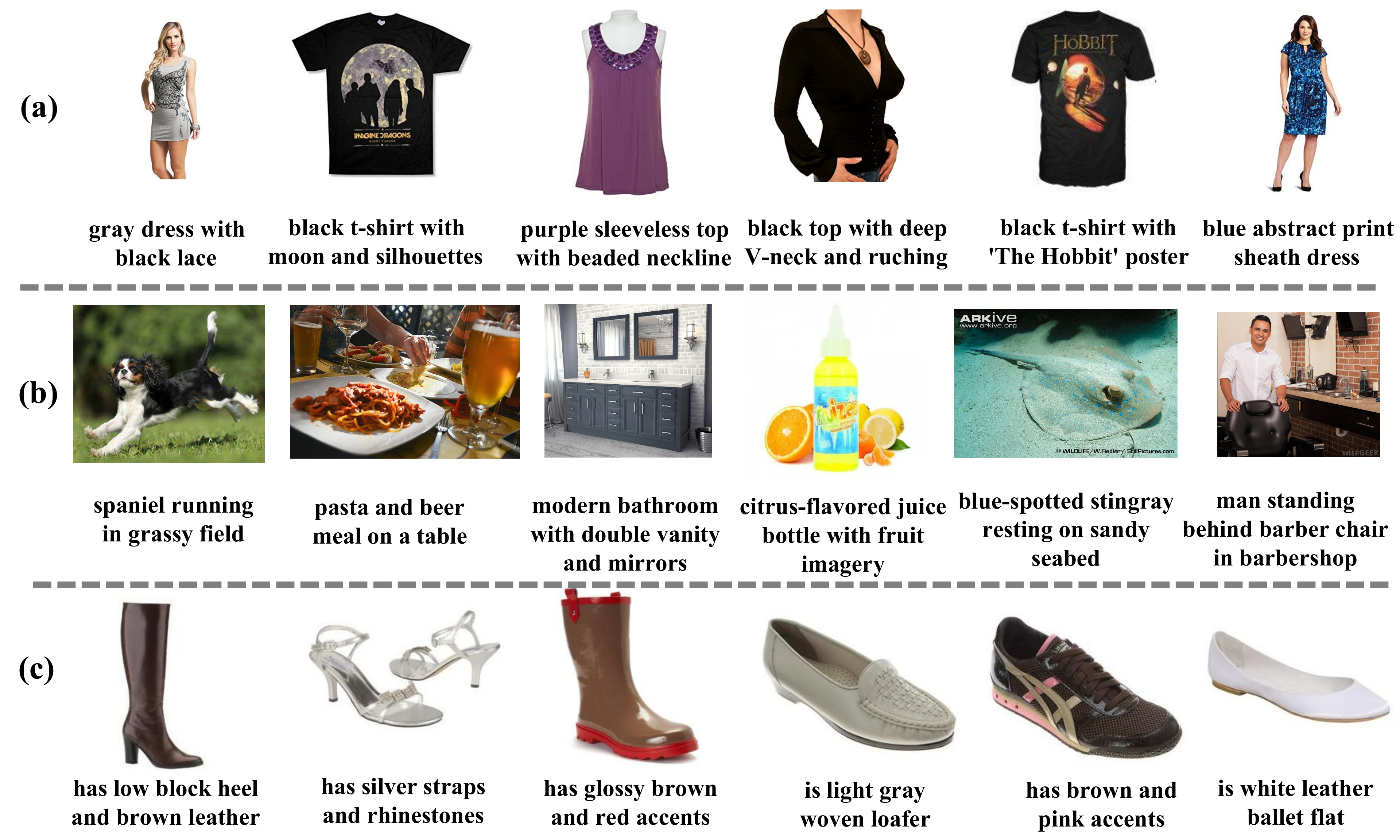}
  \vspace{-0.2cm}
  \caption{Visualization of our MCoT-generated captions for target images.}
  \vspace{-0.5cm}
  \label{fig:cot}
\end{figure*}



\section{Visualization}

\label{D}
\subsection{MCoT Cases Visualization}
\label{D1}
Fig.~\ref{fig:cot} shows examples of our MCoT-generated captions for target images across Fashion-IQ, CIRR, and Shoes datasets. Through MCoT, we successfully generate concise yet informative descriptions that accurately capture object characteristics while effectively mimicking the style of manipulation text. This approach ensures that generated captions maintain format consistency with query text and prevent the introduction of misleading or extraneous content.

\subsection{T-SNE Visualization on Fashion-IQ dataset}
\label{D2}
To validate that our method effectively mitigates the representation space fragmentation problem, we visualize query-target feature distributions on the Fashion-IQ dataset using t-SNE. As shown in Fig.~\ref{fig:tsne_2} a-b, directly evidencing the space fragmentation caused by heterogeneous modalities and asymmetric encoders. In contrast, our method achieves significantly tighter clustering both before and after training (Fig.~\ref{fig:tsne_2} c-d), demonstrating that introducing target captions establishes modal symmetry and bridges the modality gap. Moreover, query and target embeddings are positioned substantially closer under our symmetric architecture compared to SPRC, confirming superior alignment through consistent feature representations enabled by the shared-parameter Q-Former.

\end{document}